\documentclass[sigconf,nonacm]{acmart}

\AtBeginDocument{%
  }

\setcopyright{rightsretained}
\copyrightyear{2024}
\acmYear{2024}
\acmConference{SA Posters '24}{December 03-06, 2024}{Tokyo, Japan}
\acmBooktitle{SIGGRAPH Asia 2024 Posters (SA Posters '24), December 03-06, 2024}
\acmDOI{10.1145/3681756.3697906}
\acmISBN{979-8-4007-1138-1/24/12}

\acmSubmissionID{pos\textunderscore217}

\setcitestyle{square}

\usepackage{multirow}
\usepackage{bm}

\graphicspath{{figs}}

\newcommand{\cellc}[1]{\multicolumn{1}{c}{#1}}
\newcommand{\cellcl}[1]{\multicolumn{1}{c|}{#1}}
\newcommand{\cellbar}{\cellc{$-$}}

\begin{document}

\title{MambaPainter: Neural Stroke-Based Rendering in a Single Step}

\author{Tomoya Sawada}
\email{sawada@mm.doshisha.ac.jp}
\orcid{0000-0002-0391-4014}
\affiliation{%
  \institution{Doshisha University}
  \city{Kyotanabe}
  \state{Kyoto}
  \country{Japan}
}

\author{Marie Katsurai}
\email{katsurai@mm.doshisha.ac.jp}
\orcid{0000-0003-4899-2427}
\affiliation{%
  \institution{Doshisha University}
  \city{Kyotanabe}
  \state{Kyoto}
  \country{Japan}
}

\renewcommand{\shortauthors}{Tomoya Sawada and Marie Katsurai}

\begin{abstract}
Stroke-based rendering aims to reconstruct an input image into an oil painting style by predicting brush stroke sequences.
Conventional methods perform this prediction stroke-by-stroke or require multiple inference steps due to the limitations of a predictable number of strokes.
This procedure leads to inefficient translation speed, limiting their practicality.
In this study, we propose MambaPainter, capable of predicting a sequence of over 100 brush strokes in a single inference step, resulting in rapid translation.
We achieve this sequence prediction by incorporating the selective state-space model.
Additionally, we introduce a simple extension to patch-based rendering, which we use to translate high-resolution images, improving the visual quality with a minimal increase in computational cost.
Experimental results demonstrate that MambaPainter can efficiently translate inputs to oil painting-style images compared to state-of-the-art methods.
The codes are available at \href{https://github.com/STomoya/MambaPainter}{this URL}.
\end{abstract}

\ccsdesc[500]{Computing methodologies~Image processing}
\ccsdesc[500]{Computing methodologies~Neural networks}

\keywords{State-space models, stroke-based rendering, neural style transfer}
\begin{teaserfigure}
  \centering
  \includegraphics[width=\linewidth]{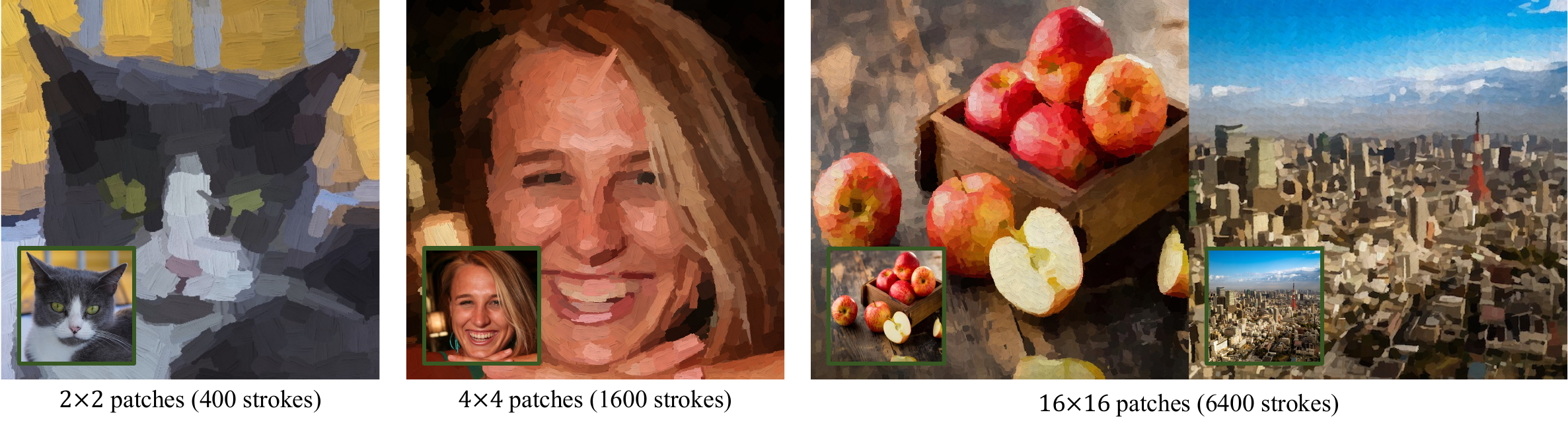}
  \caption{Translation results of MambaPainter with different number of patches. Our model can predict over 100 brush stroke parameters with a single inference, enabling efficient translation with superior visual appearance. Zoom in for best view.}
  \Description{}
  \label{fig:teaser}
\end{teaserfigure}


\maketitle

\section{Introduction}

Stroke-based rendering (SBR) aims to reconstruct a source image into an oil painting style by sequentially placing brush strokes on a canvas.
Traditional studies predict this sequence of brush strokes based on  multiple inference steps, resulting in slow generation speed and limited their practicality.
Deep learning (DL)-based approaches address this limitation by predicting a set of brush strokes using a feed-forward network, thus reducing the number of required inference steps.
However, the number of strokes that can be predicted remains limited, leaving room for achieving faster translation.
In this study, we propose MambaPainter, which is capable of predicting a sequence of over 100 brush strokes in a single inference step, enabling the efficient translation of source images to an oil painting style.
Our model is based on the selective state-space model (SSM)~\cite{mamba_gu_2024}, also known as Mamba, which has been demonstrated to model long sequences efficiently.
Our experiments present MambaPainter's superior inference efficiency on multiple image resolutions compared to state-of-the-art (SOTA) methods.
The codes are available at \url{https://github.com/STomoya/MambaPainter}.





\begin{table*}[t]
    \centering
    \small
    \caption{Efficiency analysis and quantitative evaluation results. The quantitative metrics were calculated on the $512\times512$ pixel results. Note that CNP resized the source images to $512\times512$ pixels and Im2Oil can be applied to arbitrary resolutions.}
    \label{tab:duration}
    \begin{tabular}{l|c|r|rrr|rr}
    \toprule
      \multirow{2}{*}{Method}  &  \multirow{2}{*}{\# strokes / step}  &  \multirow{2}{*}{\# params.}  &  \multicolumn{3}{c|}{time (sec.)}  &  \cellc{\multirow{2}{*}{$\mathcal{L}_{MSE}\downarrow$}}  &  \cellc{\multirow{2}{*}{LPIPS$\downarrow$}}  \\
      &&&  \cellc{$256\times256$}  &  \cellc{$512\times512$}  &  \cellcl{$1024\times1024$}  &&  \\
    \midrule
      Im2Oil~\cite{im2oil_tong_2022}  &  $-$  &  \cellcl{$-$}  &  \multicolumn{3}{c|}{$83.177\pm28.553$}  &  $0.0098\pm0.0071$  &  $0.1981\pm0.0626$  \\
      PT~\cite{pt_liu_2021} &  $0-8$  &  9.1M  &  $0.102\pm0.002$  &  $0.388\pm0.012$  &  $1.523\pm0.058$  &  $0.0122\pm0.0072$  &  $0.2898\pm0.0706$  \\
      CNP~\cite{cnp_hu_2023} &  5  &  11.2M  &  \cellbar  &  $13.372\pm0.196$  &  \cellcl{$-$}  &  $\bm{0.0056\pm0.0069}$  &  $0.1837\pm0.0836$  \\
    \midrule
      MambaPainter (Ours)  &  100  &  12.9M  &  $\bm{0.073\pm0.005}$  &  $\bm{0.198\pm0.007}$  &  $\bm{0.851\pm0.005}$  &  $0.0057\pm0.0033$  &  $\bm{0.1604\pm0.0531}$  \\
    \bottomrule
    \end{tabular}
\end{table*}

\section{MambaPainter}



\begin{figure}
    \centering
    \small
    \includegraphics[clip,width=0.95\linewidth]{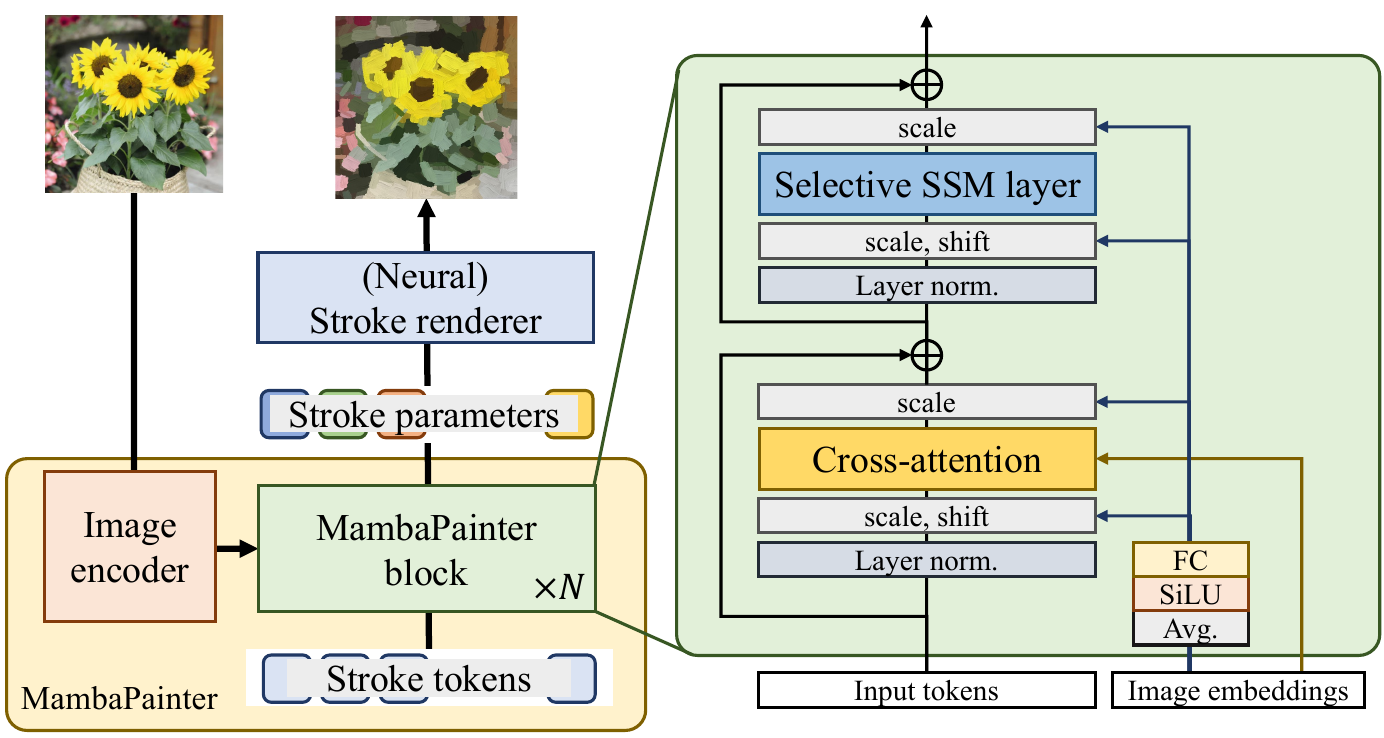}
    \caption{Overview of MambaPainter. Notably, the stroke renderer can either be a neural or non-differentiable renderer at inference time.}
    \label{fig:architecture}
    \Description{}
    
\end{figure}

Figure~\ref{fig:architecture} shows the architecture of MambaPainter.
To train our model, we utilized a neural network as a differentiable stroke renderer.
This neural stroke renderer accepts stroke parameters as inputs and outputs a transparent grey-scale image representing the brush strokes.
We used the same architecture as~\cite{cnp_hu_2023}, except we inserted instance normalization after the linear layers to improve training stability.
The objectives of the neural renderer included the $L_2$ and the LPIPS loss for the color channel and the binary cross-entropy loss for the alpha channel, resulting in brush stroke images with sharp edges and textures.




MambaPainter contains two components: a stroke predictor and an image encoder.
The former predicts a sequence of brush stroke parameters that can reconstruct the input source image.
This stroke predictor consists of selective SSM layers and cross-attention layers stacked in a interleaving manner.
By incorporating selective SSM layers, which have shown strong performance in modeling long sequences, MambaPainter can predict a more significant number of strokes compared to conventional methods.
In our study, we used a slimmed VMamba~\cite{vmamba_liu_2024} as the image encoder, considering the affinity with the stroke predictor.
We first trained MambaPainter using only $L_2$ loss.
Then, to improve the diversity of the predicted brush strokes, we added the non-saturating variant of the generative adversarial network objective after the first half of the training.
We used the ImageNet dataset as the training data for our model.



After training, the model can be applied to high-resolution images using a simple patch-based rendering approach.
In previous studies, the source image was divided into multiple patches, which were translated in a parallel manner and then merged into a single image.
However, as pointed out in~\cite{cnp_hu_2023}, this patch-based approach causes discontinuity at the edges of the patches after merging.
This problem can be attributed to the use of non-overlapping patches.
The main problem with overlapping patches is that the order of the strokes will not be preserved near the patch's edges, thus we created intermediate canvases for every 10 strokes.
Then, we alpha-blended these intermediates into a single image.
This simple extension addresses the discontinuities near the patch edges with a minimal increase in computational cost.

\section{Evaluation}


We compared our method with SOTA DL-based methods, namely compositional neural painter (CNP)~\cite{cnp_hu_2023}, PaintTransformer (PT)~\cite{pt_liu_2021}, and a search-based method Im2Oil~\cite{im2oil_tong_2022}.
We used 1,000 images from the validation set of the ImageNet dataset and measured the inference duration.
We performed this inference on three resolutions except Im2Oil, which can translate images of arbitrary resolutions.
We calculated the L2 distance and LPIPS between the original images and the translation results on $512\times512$ pixels as a quantitative metric, following~\cite{cnp_hu_2023}.

Table~\ref{tab:duration} presents the inference efficiency and quantitative evaluation scores.
In all image resolutions, MambaPainter was capable of translating images more efficiently compared to conventional methods.
Furthermore, our method showed the best performance on the LPIPS metric and the second-best score on L2 distance compared to the other methods.
Figure~\ref{fig:teaser} shows examples of the translated results.
MambaPainter could adequately translate the input images to an oil-painting style using only a single inference step.

\section{Conclusion and Future Works}

In this study, we introduced MambaPainter, which can predict a sequence of over 100 stroke parameters in a single inference step.
Our method can efficiently translate source images into an oil painting style with superior reconstruction performance compared to SOTA methods.
In future works, we will evaluate the performance of our method when applied to neural style transfer.

\bibliographystyle{ACM-Reference-Format}
\bibliography{main}


\newpage

\appendix

\setcounter{page}{1}

\section{Selective SSM}

SSMs are sequence models that hove shown strong performance on long-range dependency with linear scaling memory efficiency w.r.t the sequence length.
In general form, a linear SSM can be formulated as follows:

\begin{equation}
\label{eq:ssm}
    \begin{split}
        x'(t) &= \bm{A}x(t) + \bm{B}u(t), \\
        y(t) &= \bm{C}x(t) + \bm{D}u(t).
    \end{split}
\end{equation}
In mapping an input stimulation $u(t)$ to response $y(t)$ through an implicit latent state sequence $x'(t)$.
$\bm{A}$, $\bm{B}$, $\bm{C}$, and $\bm{D}$ are learnable weighting parameters.

To integrate SSMs to deep-learning models, a timescale parameter $\bm{\Delta}$ is used to transform the continuous parameters $\bm{A}, \bm{B}$ to discrete parameters $\overline{\bm{A}}, \overline{\bm{B}}$, respectively.
The zero-order hold is commonly used to perform the discretization as follows:

\begin{equation}
    \begin{split}
        \overline{\bm{A}} &= \exp(\bm{\Delta}\bm{A}), \\
        \overline{\bm{B}} &= (\bm{\Delta}\bm{A})^{-1} (\exp(\bm{\Delta}\bm{A}) - \bm{I}).
    \end{split}
\end{equation}
Then, Eq. (\ref{eq:ssm}) can be rewritten in the following discretized form:

\begin{equation}
    \begin{split}
        x_t &= \overline{\bm{A}}x_{t-1} + \overline{\bm{B}}u_t, \\
        y_t &= \bm{C}x_t + \bm{D}u_t.
    \end{split}
\end{equation}

Eq. (\ref{eq:ssm}) is linear time invariant, which indicates that all data points in the sequence are treated equally.
Selective SSMs, also known as Mamba, addressed this limitation by introducing an input-dependent selection mechanism to emphasize or forget elements in the input sequence.

\section{Additional results}

\begin{figure}[t]
    \centering
    \includegraphics[clip,width=\linewidth]{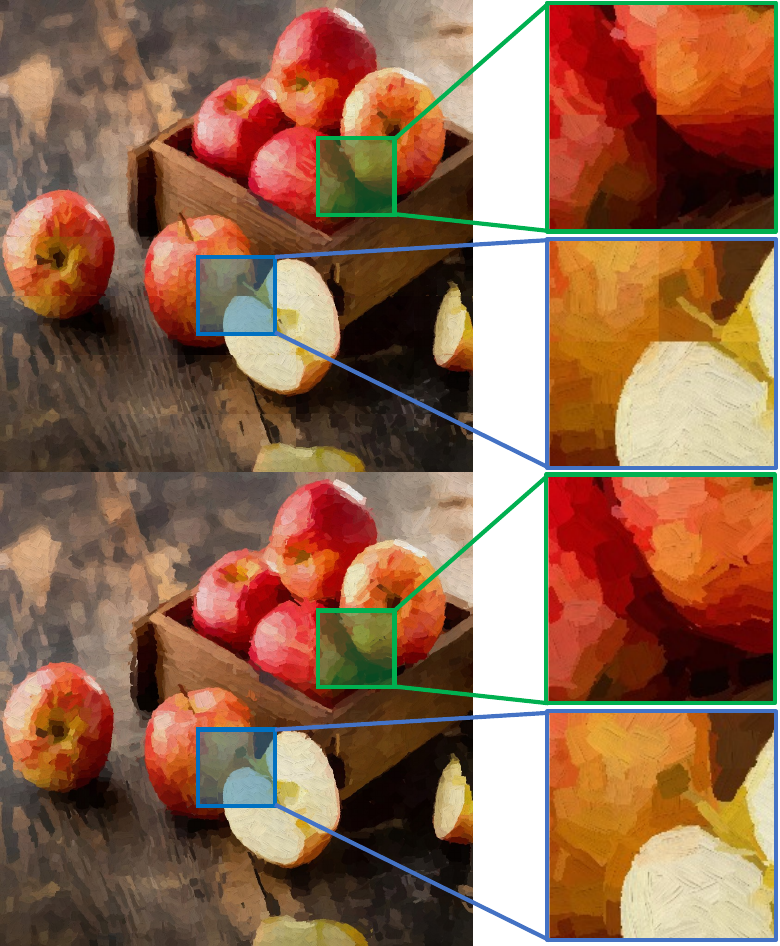}
    \caption{Comparison of translation results using non-overlapping patches (top) and overlapping patches (bottom).}
    \label{fig:patches}
\Description{}
\end{figure}

\begin{figure*}[t]
    \centering
    \includegraphics[clip,width=0.84\linewidth]{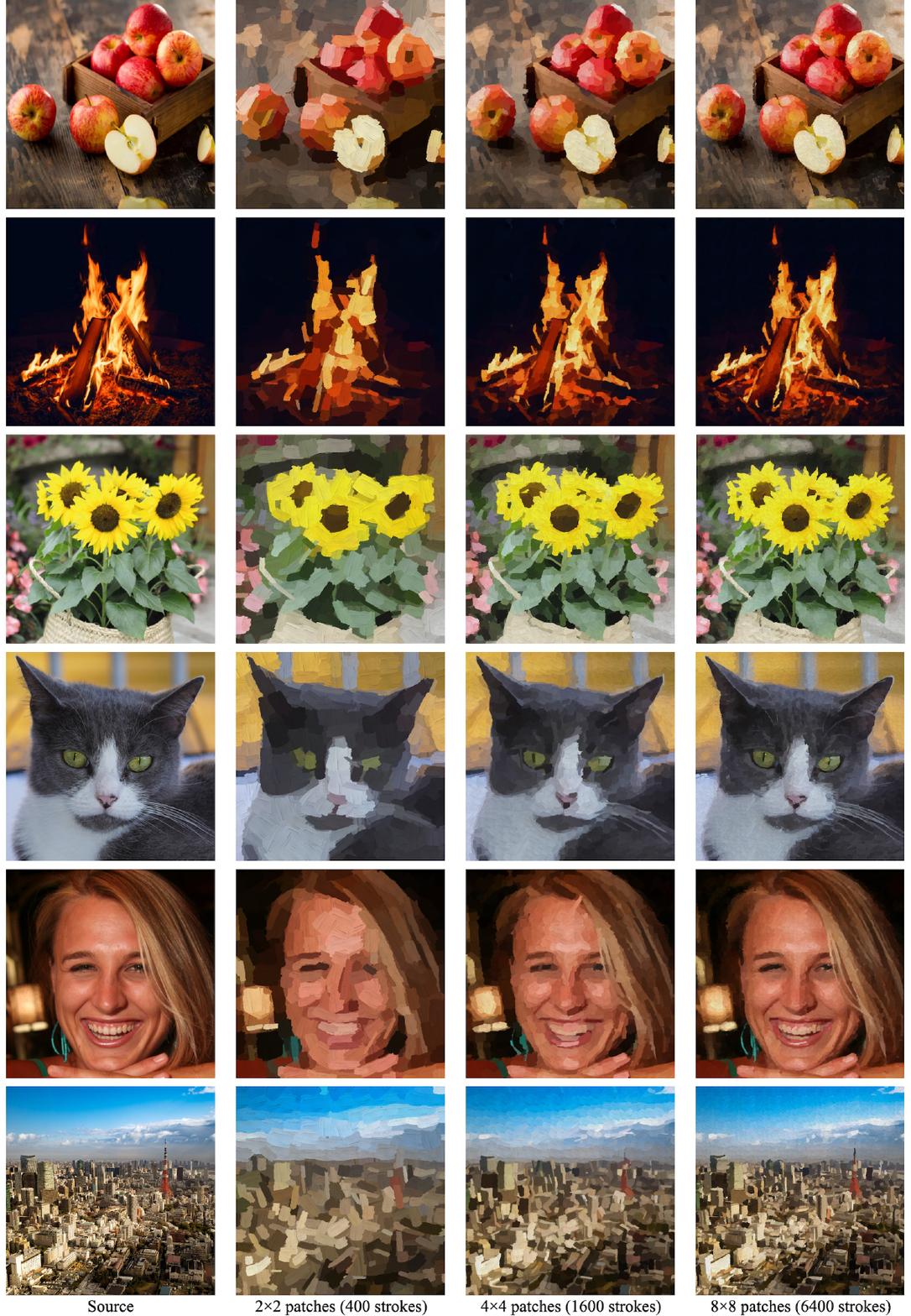}
    \caption{Additional translation results with different number of patches.}
    \label{fig:additional}
\Description{}
\end{figure*}

Figure~\ref{fig:patches} compares the translation results between overlapping and non-overlapping patches.
Our proposed overlapping patch-based approach was able to translate the source image, maintaining continuity around the patches' edges.
Figure~\ref{fig:additional} presents additional translation results using various numbers of patches.

\clearpage

\end{document}